%% file: iclr2026_conference.tex
\documentclass{article} 
\usepackage{iclr2026_conference,times}
\usepackage{amsmath,amsfonts,bm}

\input{math_commands.tex}

\usepackage{hyperref}
\usepackage{url}
\usepackage{subfigure}
\usepackage{subcaption}
\usepackage{graphicx}
\usepackage{listings}
\usepackage{xspace}
\usepackage{enumitem}
\usepackage{subcaption}
\usepackage{xspace}

\setlist[enumerate]{topsep=0pt, partopsep=0pt, parsep=0pt, itemsep=0pt}
\setlist[itemize]{topsep=0pt, partopsep=0pt, parsep=0pt, itemsep=0pt}

\lstdefinelanguage{pddl}{
  keywords={define, domain, problem, types, predicates, action, 
            parameters, precondition, effect, objects, init, goal, 
            and, or, not, when, forall, exists},
  sensitive=true,
  comment=[l]{;},
  morecomment=[s]{(:}{)},
  moredelim=[is][\color{blue}]{?}{\ },
}

\usepackage{booktabs, multirow, tabularx, siunitx, makecell}
\newcolumntype{Y}{>{\raggedright\arraybackslash}X} 

\newcommand{\vaf}{\textsc{vlm-as-formalizer}\xspace}
\newcommand{\pddl}[1]{\textsc{#1}\xspace}

\newcommand{\penn}{%
  \hspace{1pt}
  \begingroup\normalfont
  \includegraphics[height=1.3\fontcharht\font`\B]{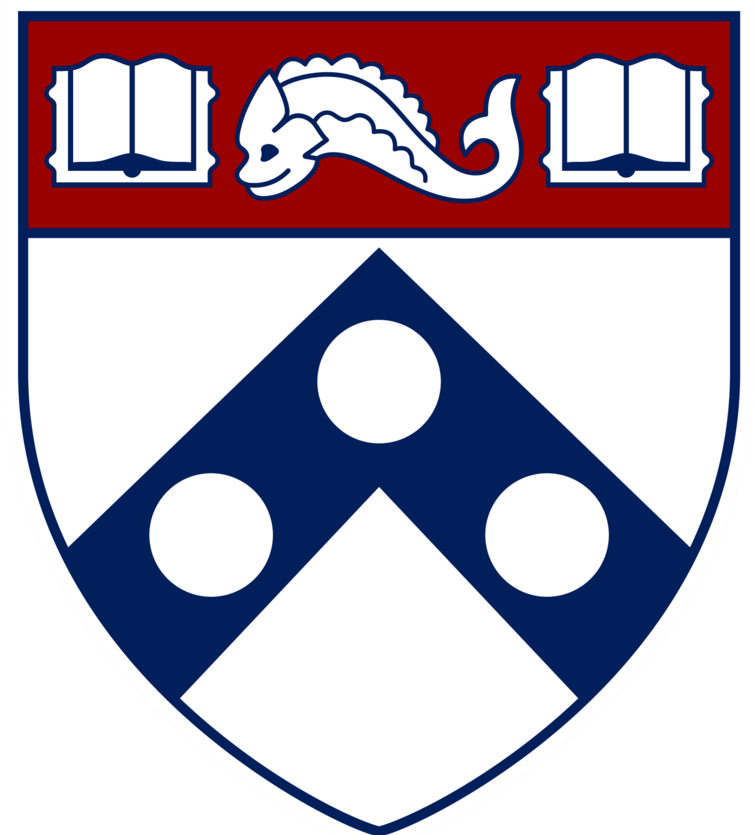}%
  \endgroup
  \hspace{1pt}
}
\newcommand{\drexel}{%
  \hspace{1pt}
  \begingroup\normalfont
  \includegraphics[height=1.3\fontcharht\font`\B]{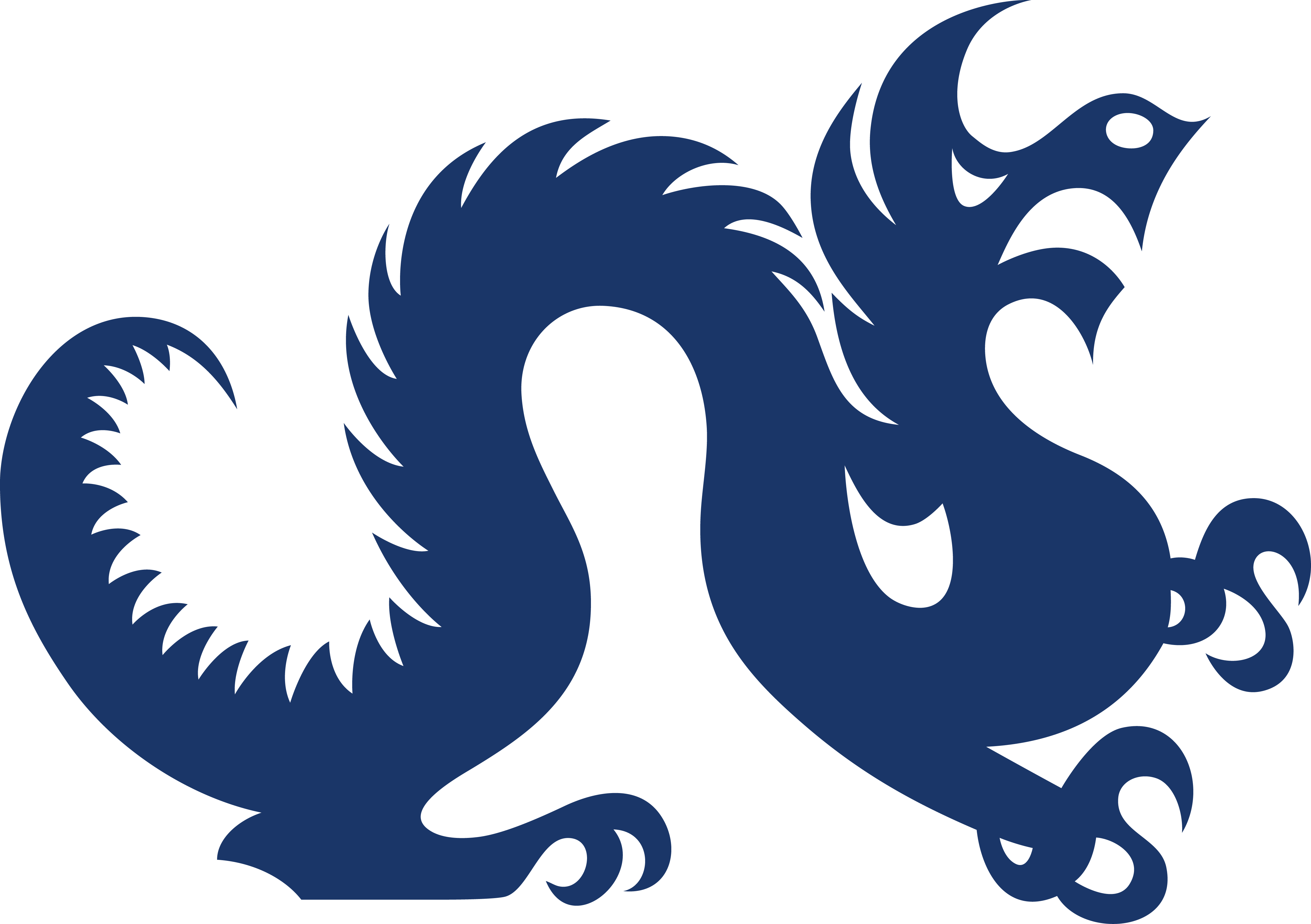}%
  \endgroup
  \hspace{1pt}
}
\newcommand{\jhu}{%
  \hspace{1pt}
  \begingroup\normalfont
  \includegraphics[height=1.3\fontcharht\font`\B]{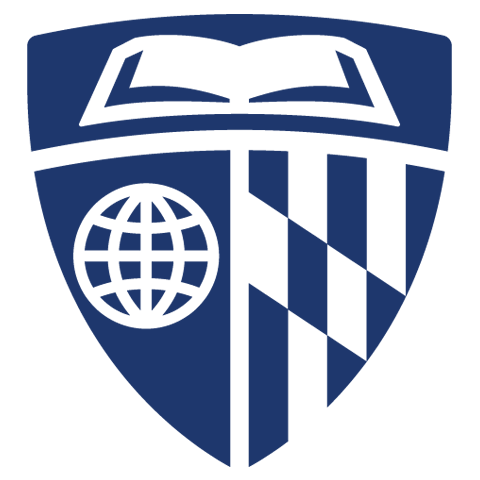}%
  \endgroup
  \hspace{1pt}
}

\title{Vision Language Models Cannot Plan, \\but Can They Formalize?}

\author{Muyu He\drexel\thanks{Work done as an intern at Drexel University.} \quad Yuxi Zheng\drexel \quad Yuchen Liu\drexel\\ Zijian An\drexel \quad Bill Cai\drexel \quad Jiani Huang\penn \\ Lifeng Zhou\drexel \quad Feng Liu\drexel \quad Ziyang Li\jhu \quad Li Zhang\drexel \\
  \drexel Drexel University \hspace{4pt} \penn University of Pennsylvania \hspace{4pt} \jhu Johns Hopkins University \\
  {\tt muyuhe@seas.upenn.edu} \quad {\tt hz466@drexel.edu}
}

\iclrfinalcopy 
\begin{document}

\maketitle

\begin{abstract}
The advancement of vision language models (VLMs) has empowered embodied agents to accomplish simple multimodal planning tasks, but not long-horizon ones requiring long sequences of actions.
In text-only simulations, long-horizon planning has seen significant improvement brought by repositioning the role of LLMs. 
Instead of directly generating action sequences, LLMs translate the planning domain and problem into a formal planning language like the Planning Domain Definition Language (PDDL), which can call a formal solver to derive the plan in a verifiable manner. 
In multimodal environments, research on \vaf remains scarce, usually involving gross simplifications such as predefined object vocabulary or overly similar few-shot examples. 
In this work, we present a suite of five \vaf pipelines that tackle one-shot, open-vocabulary, and multimodal PDDL formalization. 
We evaluate those on an existing benchmark while presenting another two that for the first time account for planning with authentic, multi-view, and low-quality images. 
We conclude that \vaf greatly outperforms end-to-end plan generation. We find that visual grounding of object relations remains the primary bottleneck for weaker VLMs, while stronger models have largely overcome this limitation.
While generating intermediate, textual representations such as captions or scene graphs partially compensate for the performance, their inconsistent gain leaves headroom for future research directions on multimodal planning formalization.\footnote{Code and data are available at \url{https://github.com/RiddleHe/vlm-as-formalizer}.}
\end{abstract}

\input{sections/intro}
\input{sections/problem_definition}
\input{sections/benchmark}
\input{sections/experimental_setup}
\input{sections/empirical_findings}

\input{sections/related_work}
\input{sections/conclusion}


\bibliography{iclr2026_conference}
\bibliographystyle{iclr2026_conference}


\end{document}

%% file: math_commands.tex
\newcommand{\pipeDP}{\textsc{Direct-P}\xspace}
\newcommand{\pipeCP}{\textsc{Caption-P}\xspace}
\newcommand{\pipeSGP}{\textsc{SG-P}\xspace}
\newcommand{\pipeASGP}{\textsc{AP-SG-P}\xspace}
\newcommand{\pipeESGP}{\textsc{EP-SG-P}\xspace}
\newcommand{\pipeDPP}{\textsc{Direct-Plan}\xspace}

\def\eqref#1{equation~\ref{#1}}

\def\1{\bm{1}}

\DeclareMathAlphabet{\mathsfit}{\encodingdefault}{\sfdefault}{m}{sl}
\SetMathAlphabet{\mathsfit}{bold}{\encodingdefault}{\sfdefault}{bx}{n}



%% file: sections/intro.tex
\section{Introduction}

Embodied planning has seen impressive advances in the last few years, especially with the rise of vision language models (VLM)  and vision language action models (VLA).
The standard setup involves giving the model interleaved vision-language inputs, such as images or videos and a natural language instruction, and expecting the model to predict a sequence of actions that logically form a plan to achieve the specified goal.
While either VLM or VLA models may directly output high-level or low-level actions, their performance is limited in long-horizontal planning with little to no interpretability \citep{liu2023egocentric,yang2025lohovlaunifiedvisionlanguageactionmodel}.  A mainstream alternative is a hierarchical pipelining approach that involves multiple models for detecting objects, extracting relations, predicting task-level actions, and translating to motion-level actions \citep{yenamandra2023homerobot,10801328}. While modular, such approach requires training for each modules in specific domains, thus lacking few-shot generalization abilities in an open-vocabulary setting. 

Large language models (LLM), the basis of VLM and VLA, have similarly driven great progress in textual planning with two leading paradigms under few-shot settings. Given a textual description of the environment and the goal, LLM-as-planner directly generates the actions \citep{wei-etal-2025-plangenllms}. In addition to mixed performance, this approach offers little interpretability and verifiability. Alternatively, LLM-as-formalizer instead generates a formal language like the Planning Domain Definition Language (PDDL) which can be input into a symbolic solver to derive a plan deterministically \citep{tantakoun-etal-2025-llms}. Powered by pre-trained LLM's strong in-context learning skills and code generation ability, this approach has shown promising performance while offering some formal guarantee that is crucial for high-stakes domains. Despite the emerging success of LLM-as-formalizer, application to VLM has been understudied in multimodal planning environments, with some incomplete attempts leveraging non-visual cues \citep{li2025bilevellearningbilevelplanning, kwon2025fastaccuratetaskplanning} or close-vocabulary, few-shot examples \citep{herzog2025domainconditionedscenegraphsstategrounded}. 
As a result, such explorations are limited to particular tasks and lack generality \citep{jenamani2025feastflexiblemealtimeassistanceinthewild}.

We advance \vaf as an effective paradigm on long-horizon, one-shot, open-vocabulary, visual-language planning tasks. We are the first to systematically evaluate its strengths and weaknesses. To do so, we consider 2 VLMs and design 5 \vaf pipelines including generating a detailed caption or scene graph as an intermediate step to PDDL formulation, compared with one VLMs-as-planner baseline. 
We evaluate each method over three vision-based datasets that drastically differ in difficulty and realism. 
On top of one existing visual Blocksworld dataset by \cite{shirai2024visionlanguageinterpreterrobottask} that is small-scale, simulated, and fully observable via one single image, we propose a novel challenge dataset, \pddl{Blocksworld-Real}, based on real images captured by sensors of physical robots. For each planning problem, we provide multiple photos from ego-centric viewpoints, challenging the model to identify and track objects across perspectives. These photos also closely resemble real-world visual conditions by including occlusion, motion blur, discoloration, and noisy background. We derive a similar multi-view planning dataset from ALFRED \citep{shridhar2020alfred}, where the planning problem is described by multiple rendered images. 

Our results position \vaf as a much stronger and more generalizable paradigm than end-to-end planning for long-horizon, visual-language planning. Even so, weaker VLMs still exhibit a clear bottleneck in visual grounding of object relations, while stronger models close this gap on simpler benchmarks but continue to struggle on complex, multi-view environments. While intermediate representations such as captions or scene graphs help, their inconsistent gain suggests future efforts. 

%% file: sections/problem_definition.tex
\section{Problem Formulation}
\label{sec:problem_formulation}

\subsection{Formal Definition}

Formally, we define the \textit{Vision-PDDL-Planning} task in line with established literature of STRIPS \citep{10.5555/1622876.1622939} as follows (Figure~\ref{fig:methodology}). 
The input consists of a triplet $(V, I, \mathcal{D})$, where:
\begin{enumerate}
    \item $V = {v_1, \ldots, v_n}$ is a sequence of $n$ images, with each image $v_i$ representing (possibly partial) observations of the initial environment;
    \item $I$ is a natural language instruction specifying the goal;
    \item $\mathcal{D}$ is a PDDL domain file, which formally defines the planning environment.
\end{enumerate}
The domain file $\mathcal{D}$ provides 
the type system for entities, 
a set of relational predicates $R$, and 
a set of parameterized actions $A$. 
Each action is specified by its preconditions and effects, both expressed as conjunctive logical formulas over the predicates.

The final objective of the Vision-PDDL-Planning task is to generate a plan $L = [a_1(\bar{e}_1), \ldots, a_m(\bar{e}_m)]$, where each 
$a_i \in A$ is an action schema defined in the domain $\mathcal{D}$, and 
each $\bar{e}_i$ is a tuple of grounded entities corresponding to the parameters of $a_i$. 
The plan $L$, when executed in the environment represented by the images $V$, must achieve the goal specified by the instruction $I$.

While one could imagine an end-to-end model that maps the input triplet $(V, I, \mathcal{D})$ directly to a plan $L$, in this work we focus on a modular paradigm: \pddl{VLM-as-Formalizer}. 
Specifically, this involves first generating a PDDL problem file $\mathcal{P}$, which encodes the initial state and goal extracted from $V$ and $I$, and then employing a PDDL solver to compute a valid plan. 

At a high level, the problem file $\mathcal{P}$ is defined by three main components: 
objects ($E$), 
the initial state ($s_0$), and 
the goal state ($\psi{\texttt{goal}}$):
\begin{enumerate}
    \item Objects: $E = {e_1, \ldots, e_n}$ is the set of named entities present in the environment. Each object is an instance of an entity type defined in $\mathcal{D}$.
    \item Initial State: $s_0 \in \mathcal{S}$ is a set of relational facts of the form $r(\bar{e})$, where $r \in R$ is a relational predicate defined in $\mathcal{D}$; $\bar{e}$ is a tuple of entities from $E$ that participate in relation $r$. 
    \item Goal State: $\psi_{\texttt{goal}}: \mathcal{S} \rightarrow \mathbb{B}$ is a boolean formula over relational facts, where $S$ is the state space and $\mathbb{B} = {\text{True}, \text{False}}$. When $\psi_{\texttt{goal}}(s^*) = \text{True}$ we say that the state $s^*$ achieves the problem goal.
\end{enumerate}
Here, the state space $\mathcal{S}$ comprises all possible configurations of relational facts over the object set $E$ with the predicates $R$. 
Each state $s \in \mathcal{S}$ is a set of instantiated facts $r(\bar{e})$, describing which relations hold among the objects.
A PDDL solver takes as input the domain-problem file pair $(\mathcal{D}, \mathcal{P})$ and search for a plan. 
If a plan $L$ is found, executing it from the initial state $s_0$ yields a sequence of intermediate states $s_1, s_2, \ldots, s^*$ such that the goal formula $\psi_{\texttt{goal}}(s^*)$ evaluates to $\text{True}$.

\input{figures/methodology}

\subsection{Evaluation Metrics}

To comprehensively measure the planning ability of \vaf methods, we propose two complementary sets of metrics: task-level metrics and scene-level metrics.

\textit{Task-level} metrics evaluate the direct planning success of the method. 
Due to the nature of the Vision-PDDL-Planning task, there are three levels of success that a correct $\mathcal{P}$ needs to attain.
The basic level of success is \textbf{compilation success} which is defined as a boolean function that evaluates to True if the PDDL solver can compile $\mathcal{P}$ into runnable code.
The necessary and sufficient condition for compilation success is syntax correctness in $\mathcal{P}$.
A level above is \textbf{planner success} which is defined as a boolean function that evaluates to True if the PDDL solver can find $L$ after exhaustive search.
$\mathcal{P}$ needs to have non-contradicting initial states and goal states to achieve planner success.
Finally, the most important metric is \textbf{simulation success}, which is defined as a boolean function that evaluates to True if the found $L$ can start from the initial states in the ground truth $\mathcal{P}$ to reach the target goal states.
For each metric, we report the average success rate of all tasks in a benchmark as indicators for pipeline performance.

To provide finer-grained insights into the problem files generated by \vaf, we also consider \textit{scene-level} metrics which evaluate the VLM's representation of the objects, initial states, and goals against a ground-truth $\mathcal{P}$, reporting micro-averaged \textbf{precision}, \textbf{recall} and \textbf{F1} for each category, computed by aggregating the true positive, false positive, and false negative counts across all tasks and benchmarks.
A model with a low recall on any of the three categories will fail to find a plan, as the solver will not correctly satisfy all relevant conditions. 
A model with a high recall but a low precision might find a plan but will do so inefficiently, which is not accounted for by task-level metrics.

%% file: figures/methodology.tex
\begin{figure}
    \includegraphics[width=\linewidth]{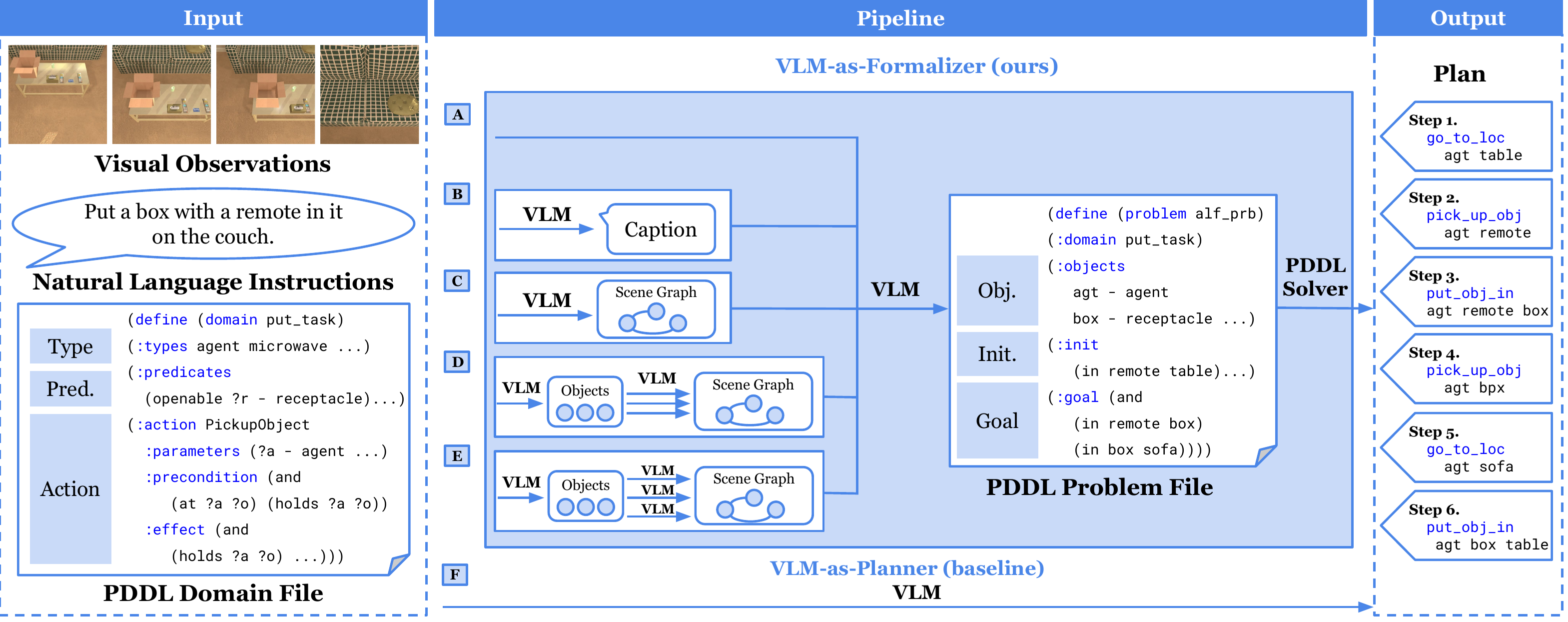}
    \caption{
        An illustration of the Vision-PDDL-planning task. The input includes visual observations of entities, natural language instructions of the goal, and a PDDL domain file. While a baseline might use a VLM to directly generate the plan, we advocate for the 5 \vaf pipelines that generates the problem file which deterministically derives the plan with a PDDL solver.
    }
    \label{fig:methodology}
\end{figure}

%% file: sections/benchmark.tex
\section{Benchmarks}
\label{sec:benchmarks}

In this section, we discuss the process to curate the benchmarks for \vaf.

\subsection{Criteria}

Before instantiating the formal definition and evaluation metrics above, we first consider the criteria of suitable benchmarks to study the effectiveness of \vaf in long-horizon, multimodal planning tasks. Minimally, they should include:
\begin{enumerate}
    \item Visual observations $V$ and natural language goal instruction $I$ as input;
    \item Ground-truth domain file $\mathcal{D}_{\texttt{domain}}$ as input, coupled with ground-truth problem file $\mathcal{P}_{\texttt{problem}}$ for evaluation.
\end{enumerate}

To maximize real-life application, ideal benchmarks should be:
\begin{enumerate}
    \item \textbf{Realistic} instead of simulated, as previous work \citep{8803821,10655420} has shown a systematic performance degradation of vision models working with photo-realistic instead of rendered images;
    \item \textbf{Noisy} instead of clean, as previous work \citep{8578445,9151047} has shown a systematic performance degradation of vision models working with poor lighting, motion blurring, discoloration, or occlusion between relevant objects.;
    \item \textbf{Multi-view} instead of single-view, as recent work \citep{liu-etal-2024-mibench,wang2025muirbench} has faced the challenge of multi-image reasoning, though not in the context of planning;
\end{enumerate}

The closest, existing benchmarks that fulfill the minimal criteria to our knowledge is the visual \pddl{Blocksworld} dataset from \citet{shirai2024vision}. However, it falls short of the real-life criteria as it is \textbf{simulated} (rendered from a game engine), \textbf{clean} (under perfect lighting condition and visibility), and \textbf{single-view} (Figure~\ref{fig:benchmarks}). Moreover, it contains 10 examples with homogeneous initial conditions, potentially leading to overestimated and high-variance evaluation results. 

\input{figures/benchmarks}

\subsection{Proposed Benchmarks}

To address this issue, we propose two new benchmarks: \pddl{Blocksworld-Real} based on a \textbf{realistic} staging of the classical Blocksworld domain \citep{ipc} and \pddl{Alfred-Multi} based on the widely used ALFRED simulated environment \citep{shridhar2020alfred} (Figure~\ref{fig:benchmarks}). Naturally, both fulfill the minimal criteria with manually annotated and curated multimodal input and PDDL. They are both \textbf{noisy} under different extent of imperfect visual conditions by design, which more closely resemble the real-world environment where the robots are deployed. They are also both \textbf{multi-view}, providing multiple images that represent the same initial state of the environment from multiple perspectives. 
As a result, the same object can occur in multiple images, which poses a challenge for the VLM to identify them correctly and consistently.
Often, the two benchmarks rather uniquely define a task environment that is \textit{partially observable} with regard to each individual image, but \textit{fully observable} only with regard to the complete set of images. 
Each image provides only a fraction of the necessary information for planning and cannot be relied on as the sole reference.
Therefore, the VLM needs to extract information necessary for planning from all images. 

We now describe both benchmarks in detail.
\pddl{Blocksworld-Real} has the classic blocksworld setup: given an initial stacking of colored blocks, a model is required to create a new stacking as instructed.
We collect a total of 102 problems using blocks of distinct colors.
For each problem, which is generated programmatically, we also annotate the ground truth problem file $\mathcal{P}_\texttt{gt}$ that precisely describes the initial states and goal states of each block. We pair each problem file $\mathcal{P}_\texttt{gt}$ with an overarching domain file $\mathcal{D}_\texttt{gt}$ to a solver to get the ground truth plan. 
On average, each plan consists of 12 steps, which implies a large search space for long-horizon planning.
To get the corresponding visual inputs of \pddl{Blocksworld-Real}, we set up the exact initial stacking of each task using real blocks in a robotics lab and use a camera on a robotic arm to perform a sweep motion around the stacked blocks.
This gives us one hundred image frames from which we select four at equal temporal intervals to obtain a set of images with diverse viewpoints.



In contrast, \pddl{Alfred-Multi} simulates an indoor household environment where a model performs everyday tasks such as putting a clock on a desk or heating a meal in a microwave.
We take 150 trajectories from the training split of the original ALFRED environment to compose our tasks. 
To obtain the ground-truth problem file $\mathcal{P}_\texttt{gt}$ and visual inputs, we reverse engineer the necessary object instances and object states that must occur in $\mathcal{P}_\texttt{gt}$ for the solver to find the exact same plan.
As the necessary objects are already included in the ground-truth plan, to find each object's true grounded predicates, we take advantage of the provided PDDL file in each task to extract the relevant lines.
Then, to get the minimally sufficient set of images taken from the environment, we feed the extracted object states into the provided simulation engine, and we collect only images that show at least one of the objects that occur in $\mathcal{P}_\texttt{gt}$.
As a result, we now have a set of images which together define a fully observable environment of all relevant objects but individually define a partially observable environment of one or more objects.
This yields on average four to six images per problem.


%% file: figures/benchmarks.tex
\begin{figure}[t!]
    \includegraphics[width=\linewidth]{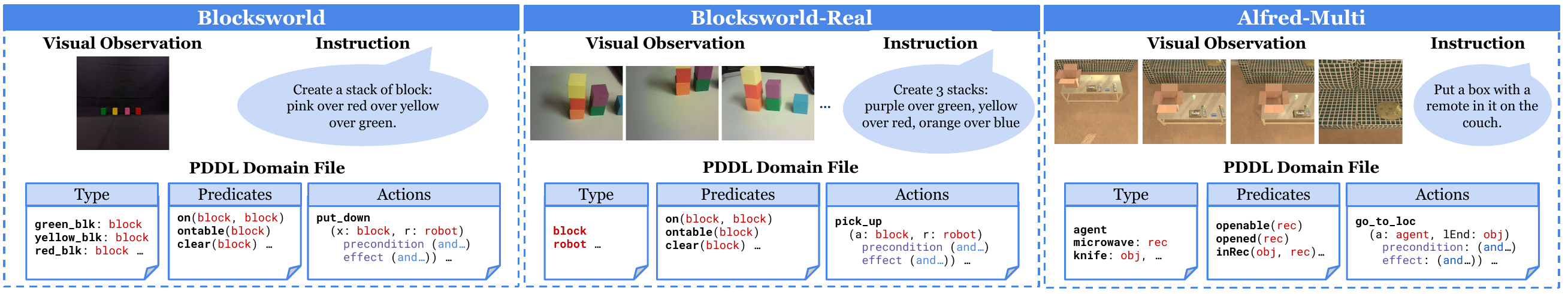}
    \caption{
        The three benchmarks in Vision-PDDL-Bench.
Each benchmark provides visual observations (single or multiple images), a natural language instruction, and a PDDL domain file as input. The task is to generate a valid plan such that, after execution, the resulting state fulfills the goal expressed in the natural language instruction.
    }
    \label{fig:benchmarks}
\end{figure}

%% file: sections/experimental_setup.tex
\section{Experimental Setup}
\label{sec:experimental_setup}

\subsection{Methodology}
\input{figures/pipeline}
As is shown in Fig.~\ref{fig:methodology}, we consider 6 pipelines that use VLMs to produce a plan for the Vision-PDDL-planning task.
Among them, the first five (A-E) fall into the category of \vaf, whereas the last (F) skips PDDL and generates the plan in an end-to-end manner.
As is discussed in Section~\ref{sec:problem_formulation}, all methods assume  a common set of inputs, which consists of a list of visual observation images and a natural language goal instruction. 

To begin with, the \pipeDP pipeline (A) directly generates $\mathcal{P}_\texttt{pred}$ in a single call to the VLM.
In the prompt, the VLM is first given an out-of-domain one-shot example of a problem file as a reference to the PDDL syntax. 
To ensure no semantic overlap with the current task, we use a toy example from the Tower of Hanoi domain.
The VLM is also allowed to output intermediate reasoning steps though we do not explicitly prompt it to do so.
We parse and extract $\mathcal{P}_\texttt{pred}$ from the output and input in into the solver to find the optimal plan.

To study the effect of scaling up test-time compute to explicitly analyze the scene, the \pipeCP pipeline (B) first generates an intermediate \textbf{scene caption} in natural language and then the problem file $\mathcal{P}_\texttt{pred}$ in the second step. 
We prompt and require the VLM to generate scene captions that consist of five aspects of the scene: (i) relevant object types and their instances, (ii) the quantity of each object type, (iii) relevant spatial relationships between the objects, (iv) task-related object properties, and (v) vision-related object properties. 
The first two aspects force the VLM to accurately enumerate the objects that will serve as grounded arguments to the initial states in $\mathcal{P}_\texttt{pred}$.
The third aspect directs the VLM to attend to the binary relations between objects in those states, whereas the fourth and fifth focuses on unary relations of each single object.
Given the scene caption in the first step, the VLM is prompted again with the same input to output the $\mathcal{P}_\texttt{pred}$.

To enforce more formality on the model generation, the \pipeSGP pipeline (C) instructs the VLM to generate a \textbf{scene graph} that describes the images.
For each object type (e.g., \texttt{block}) and each predicate (e.g., \texttt{ontable(block)}) defined in the domain file $\mathcal{D}$, the model generates all the instantiated objects (e.g., \texttt{red\_blk:block}) and grounded predicates (e.g., \texttt{ontable(red\_blk)}).
Given this scene graph, the second step is a pure translation task, where the model simply translates the instantiated objects and grounded predicates into $\mathcal{P}_\texttt{pred}$, in addition to predicting goal states based on the natural language instruction.

Concerned about possible low recalls of the grounded predicates in the scene graph generated by the previous two methods, we propose the \pipeASGP pipeline (D) that \textbf{eliminates false grounded predicates} from an exhaustive list of all possible grounded predicates. 
In the first pass, given the input, we first prompt the VLM to identify all relevant objects in the scene and their corresponding type.
We then automatically enumerate all possible grounded predicates where each argument is an object instance identified by the VLM.
In the second pass, we input \textbf{all} possible grounded predicates to the VLM at once to verify the existence each of the grounded predicates as `True' or `False' labels.
The grounded predicates labeled as `True' become the set of initial states that will go into $\mathcal{P}_\texttt{pred}$, and the identified object instances become the set of objects in $\mathcal{P}_\texttt{pred}$.
In the final pass, we send the instruction along with all identified objects and initial states to the VLM to come up with the goal states which complete the $\mathcal{P}_\texttt{pred}$. 
The model is allowed to add or remove from the previously decided initial states based on a holistic understanding of the goal not previously accessible.

Similar to the above, the \pipeESGP pipeline (E) also first enumerates all possible grounded relations.
However, to study the impact of context window size on the VLM's ability to identify the correct object relations, we iteratively pass \textbf{each} possible grounded relation, instead of all of them at once, to the VLM in a separate call. 
On our benchmarks, we find the increase in computation time is negligible when as the number of predicates is small. 

Finally, as a baseline against all five \vaf approaches, we consider \pipeDPP (F) to directly output a plan without an intermediate $\mathcal{P}_\texttt{pred}$.
Along with a one-shot out-of-domain example of the structure of the plan as a sequence of grounded actions, we instruct the model to generate a plan that consists solely of grounded actions that are defined in $\mathcal{D}$.

\subsection{Infrastructure}
We study two VLMs which are state-of-the-art in the open-source and proprietary domains, respectively.
The proprietary model we study is GPT-5.2 (G-5.2), using the OpenAI client endpoint to call the 'gpt-5.2' version of the model.
The open-source model we study is Qwen3-VL-235B (Q-235B), specifically the \texttt{Qwen/Qwen3-VL-235B-A22B-Instruct} checkpoint, which we host via vLLM on 8 NVIDIA H100 GPUs locally to enable fast inference.
We use a temperature of 0.7 for both models and use a max token count of 1024 to avoid cutoff of $\mathcal{P}_\texttt{pred}$ generation.
After the VLM predicts $\mathcal{P}_\texttt{pred}$, we pair it with the ground-truth $\mathcal{D}$ and employ the Fast Downward Planner \citep{10.5555/1622559.1622565} to deterministically find the plan.

%% file: figures/pipeline.tex
\begin{figure}
    \centering
    \includegraphics[width=\linewidth]{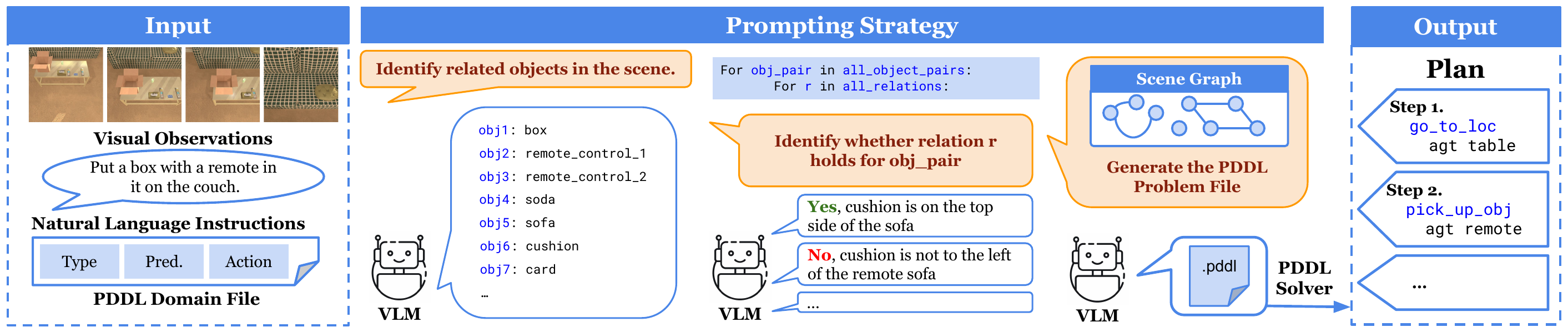}
    \caption{Example pipeline of \pipeESGP.
The prompting process involves first identifying the relevant objects, then predicting the existence of relations between them, and finally generating a PDDL problem file from the resulting scene graph. A PDDL solver is then used to produce the plan.}
    \label{fig:placeholder}
\end{figure}

%% file: sections/empirical_findings.tex
\section{Empirical Findings}
\label{sec:empirical_findings}

We report an array of constructive findings resulting from the 6 pipelines on the 3 benchmarks.

\input{figures/overall_success_rate}

\paragraph{\vaf establishes a strong advantage over end-to-end planning.} 
Across all three benchmarks and two VLMs (Figure~\ref{fig:overall_success}), \pipeDPP achieves uniformly low performance, confirming that state-of-the-art VLMs are unable to reliably generate plans in multimodal, long-horizon planning tasks. In contrast, the five \vaf pipelines consistently achieve superior performance, with the least performing pipeline \pipeESGP still gaining a considerable advantage.
The clear superiority of \vaf stands in contrast with previous work in text-based planning, where end-to-end LLM-as-planner pipelines often lead to effective results on complex domains~\citep{huang-zhang-2025-limit}. This can likely be explained by VLMs' lack of inference-time scaling of reasoning tokens as some LLMs do, which are crucial for solving such high-complexity tasks. 
Comparing the performance on our proposed \pddl{Blocksworld-Real} and the existing simulated \pddl{Blocksworld} benchmark, the added realism, multiple views, and degraded image quality add challenges to the \vaf methods, while the performance of certain pipelines remains robust. 

\paragraph{\vaf benefits from generating intermediate representations.} 
On Q-235B, \pipeCP and to a lesser extent \pipeSGP outperform \pipeDP, while on G-5.2, \pipeDP achieves comparable or superior simulation success without an intermediate step.
This pattern is mirrored in Table~\ref{tab:precision-recall}, where Q-235B benefits from intermediate representations in precision and recall while G-5.2 achieves its strongest F1 with \pipeDP directly.
This suggests that inference-time scaling via intermediate representations can compensate for weaker perceptual capacity, though stronger models may not require this step.
No advantage is observed from more complex inference techniques proposed in \pipeASGP~and \pipeESGP~.

\input{tables/precision_recall}

\paragraph{The bottleneck of \vaf is visually grounding initial states.} Recall that the objective of a \vaf pipeline on the Vision-PDDL-Planning task is to generate a problem file. While not shown in Figure~\ref{fig:overall_success}, the compilation success rate for all pipelines on all datasets is 100\%, suggesting feasibility of generating syntactically correct PDDL. Semantically, the problem file consists of three components: objects, initial states, and goal states. In our task formulation, the objects and initial states are solely informed by the visual input, while the goal states are solely inform by the textual input. 
As reported in Table~\ref{tab:precision-recall}, Q-235B shows markedly lower init-state F1 than object or goal F1, while G-5.2 achieves init-state F1 on par with or exceeding goal-state F1.
This indicates that visual grounding of object relations remains a bottleneck for weaker models, while stronger models have largely closed this gap. 

\paragraph{VLMs are more prone to omit correct states than proposing incorrect ones.}
All pipeline's predictions of objects, initial conditions, and goals show consistently lower recall than precision, highlighting VLMs’ struggles with false negatives.
The right section of Figure~\ref{fig:qualitative}, showing \pipeSGP~results, illustrates a common cause of false negatives.
In the example, to successfully predict all relevant states of a block being on the table and with no other blocks on its top, the VLM needs to predict the block \texttt{x} as being both in the state (\texttt{clear} \texttt{x}) and in (\texttt{ontable} \texttt{x}).
However, it is a common pitfall for VLMs to predict only one of the two, leading to planner failure when the PDDL solver cannot locate \texttt{x} to enable the search.

\paragraph{Intermediate representations affect how VLMs perceive the scene.}
As shown in Figure~\ref{fig:qualitative}, which compares $\mathcal{P}_{\texttt{pred}}$ generated by \pipeCP~and \pipeSGP~, the different intermediate representations induced by prompt strategies in the two pipelines lead to significant drift in the content of $\mathcal{P}_{\texttt{pred}}$. 
The prompt of \pipeCP~is \textit{structure-oriented}, guiding the VLM to first perceive objects and then capture their organization into patterns such as stacks of blocks, rather than describing objects in isolation.
This perceptual pattern is reflected in the resulting $\mathcal{P}_{\texttt{pred}}$ which accurately grounds a stack of blocks (\texttt{blue}, \texttt{orange}, \texttt{purple}) but completely mischaracterizes another stack (\texttt{red}, \texttt{yellow}).
By contrast, the prompt of \pipeSGP~is \textit{relation-centric} in the sense that it asks the VLM to iteratively ground each predicate in $\mathcal{D}_{\texttt{gt}}$, which takes multiple distinct objects as arguments.
This results in the VLM successfully grounding a complete set of relations sharing the same predicate (\texttt{ontable}) but grossly misses others (\texttt{clear}, \texttt{on}). 
At inference time, the prompting strategies of different pipelines lead to non-trivial disagreements in the VLM's perception of relational facts, causing real differences in simulation and planning success.

\paragraph{The planning superiority of \textsc{vlm-as-formalizer} is a tradeoff with token efficiency.}
We plot the number of tokens generated by each pipeline averaged across benchmarks in Figure~\ref{fig:success_rate_per_token}, coupled with the average success rate per token, calculated by dividing the average success rate by the number of total tokens in a single task. 
We witness that \pipeDPP significantly outperforms all \vaf methods in terms of token efficiency, the reverse of what we observed when judging success rates.
Among \vaf pipelines, \pipeCP, often the leading method in terms of planning success, ranks among the least token-efficient approaches.
Although it enjoys model-dependent gains of 0\% to 24\% in simulation success over \pipeDP, it also consumes more than 102\% of the total tokens through verbalization of a set of perceptual tasks.
The pronounced differences between pipelines point to a frontier for exploring improvements in VLMs’ inherent formalizer capacity, with \pipeDP striking a balance under current VLM capabilities.








\input{figures/success_rate_per_dollar}

%% file: figures/overall_success_rate.tex
\begin{figure}[t]
    \centering
    \includegraphics[width=\linewidth]{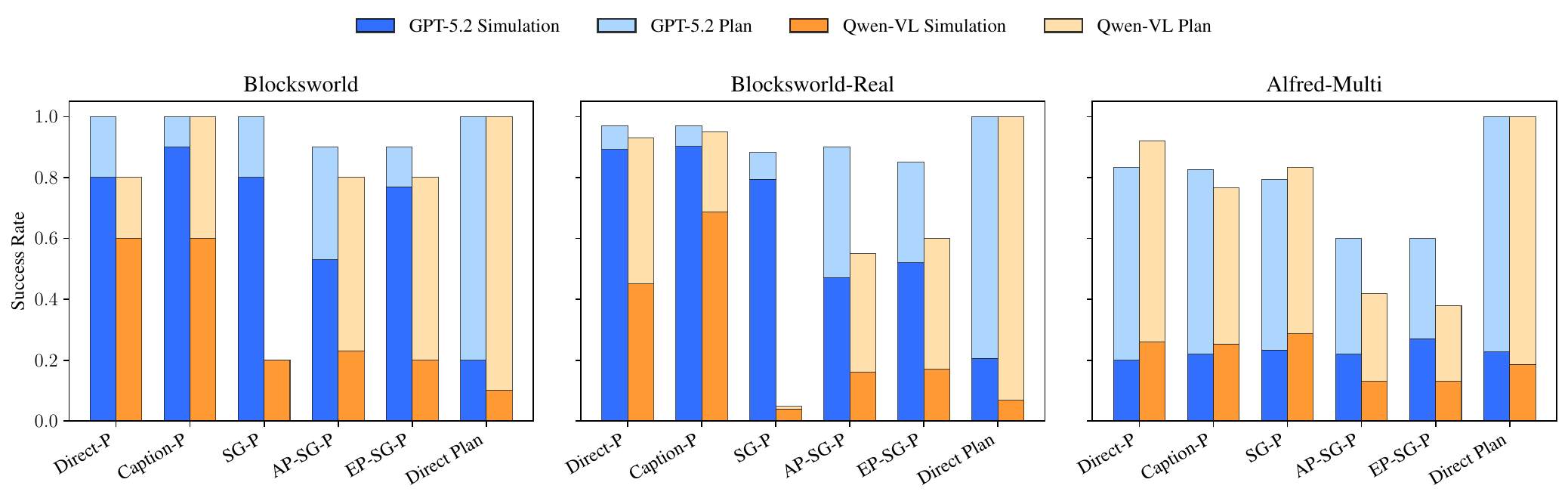}
    \caption{Success rates of six strategies across three benchmarks.
Bars represent planner success (light), and simulator success (dark), reflecting increasing levels of strictness from syntax validity to plan generation to goal-achieving execution.}
    \label{fig:overall_success}
\end{figure}

%% file: tables/precision_recall.tex
\begin{table}[t]
\centering
\caption{
We assess the quality of generated PDDL problem files along three dimensions: correctly identified objects, correctly predicted initial states, and correctly specified goal states.
We report micro-averaged Precision (\textbf{P}), Recall (\textbf{R}), and \textbf{F1} across all tasks and benchmarks for each \textsc{VLM-as-Formalizer} pipeline.}
\label{tab:precision-recall}
\scriptsize
\setlength{\tabcolsep}{2pt}
\begin{tabularx}{\textwidth}{@{} l Y *{9}{S[table-format=1.4]} @{}}
\toprule
\multicolumn{2}{c}{} &
\multicolumn{3}{c}{\textbf{Objects}} &
\multicolumn{3}{c}{\textbf{Initial States}} &
\multicolumn{3}{c}{\textbf{Goal States}} \\
\cmidrule(lr){3-5} \cmidrule(lr){6-8} \cmidrule(lr){9-11}
\textbf{Model} & \textbf{Pipeline} &
\textbf{P} & \textbf{R} & \textbf{F1} &
\textbf{P} & \textbf{R} & \textbf{F1} &
\textbf{P} & \textbf{R} & \textbf{F1} \\
\midrule
\multirow{5}{*}{\textbf{G-5.2}}
& \pipeDP & 1.0000 & \underline{\textbf{0.7475}} & \underline{\textbf{0.8555}} & 0.8840 & \underline{\textbf{0.8228}} & \underline{\textbf{0.8523}} & 0.8617 & \underline{\textbf{0.7817}} & 0.8197 \\
& \pipeCP   & 1.0000 & 0.7297 & 0.8437 & 0.8904 & 0.7988 & 0.8421 & 0.8982 & 0.7544 & \underline{\textbf{0.8201}} \\
& \pipeSGP & 1.0000 & 0.6640 & 0.7981 & \underline{\textbf{0.9186}} & 0.7426 & 0.8213 & \underline{\textbf{0.9157}} & 0.6486 & 0.7593 \\
& \pipeASGP  & 1.0000 & 0.6018 & 0.7514 & 0.7920 & 0.4064 & 0.5372 & 0.7469 & 0.6018 & 0.6665 \\
& \pipeESGP & 1.0000 & 0.6056 & 0.7544 & 0.7316 & 0.4191 & 0.5329 & 0.7308 & 0.5930 & 0.6547 \\
\midrule
\multirow{5}{*}{\textbf{Q-235B}}
& \pipeDP & 1.0000 & \underline{\textbf{0.7605}} & \underline{\textbf{0.8640}} & 0.6567 & 0.6820 & 0.6691 & 0.6887 & \underline{\textbf{0.7626}} & \underline{\textbf{0.7238}} \\
& \pipeCP & 1.0000 & 0.7203 & 0.8374 & \underline{\textbf{0.8317}} & \underline{\textbf{0.7502}} & \underline{\textbf{0.7888}} & 0.6220 & 0.7025 & 0.6598 \\
& \pipeSGP & 1.0000 & 0.2717 & 0.4272 & 0.2959 & 0.0584 & 0.0976 & \underline{\textbf{0.8075}} & 0.2049 & 0.3268 \\
& \pipeASGP  & 0.4485 & 0.2676 & 0.3352 & 0.3754 & 0.2138 & 0.2724 & 0.3991 & 0.2610 & 0.3156 \\
& \pipeESGP & 0.6250 & 0.2145 & 0.3193 & 0.4191 & 0.1323 & 0.2012 & 0.5576 & 0.2050 & 0.2997 \\
\bottomrule
\end{tabularx}
\end{table}

%% file: figures/success_rate_per_dollar.tex
\begin{figure}[t]
    \centering
    \begin{minipage}{0.47\linewidth}
        \centering
        \includegraphics[width=\linewidth]{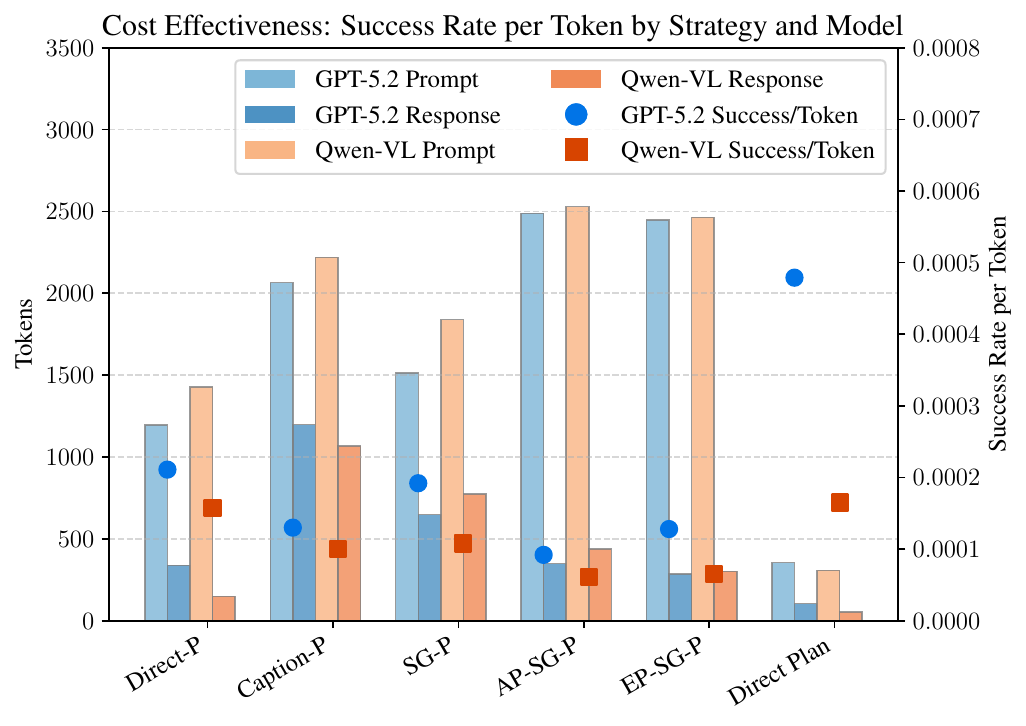}
        \caption{Cost effectiveness of LLM-based planning strategies. The stacked bars report the average number of prompt and response tokens consumed by GPT-5.2 and Qwen3-VL under each strategy. Overlaid markers indicate the simulation success rate normalized by the amount of tokens.}
        \label{fig:success_rate_per_token}
    \end{minipage}%
    \hfill
    \begin{minipage}{0.49\linewidth}
        \centering
        \includegraphics[width=\linewidth]{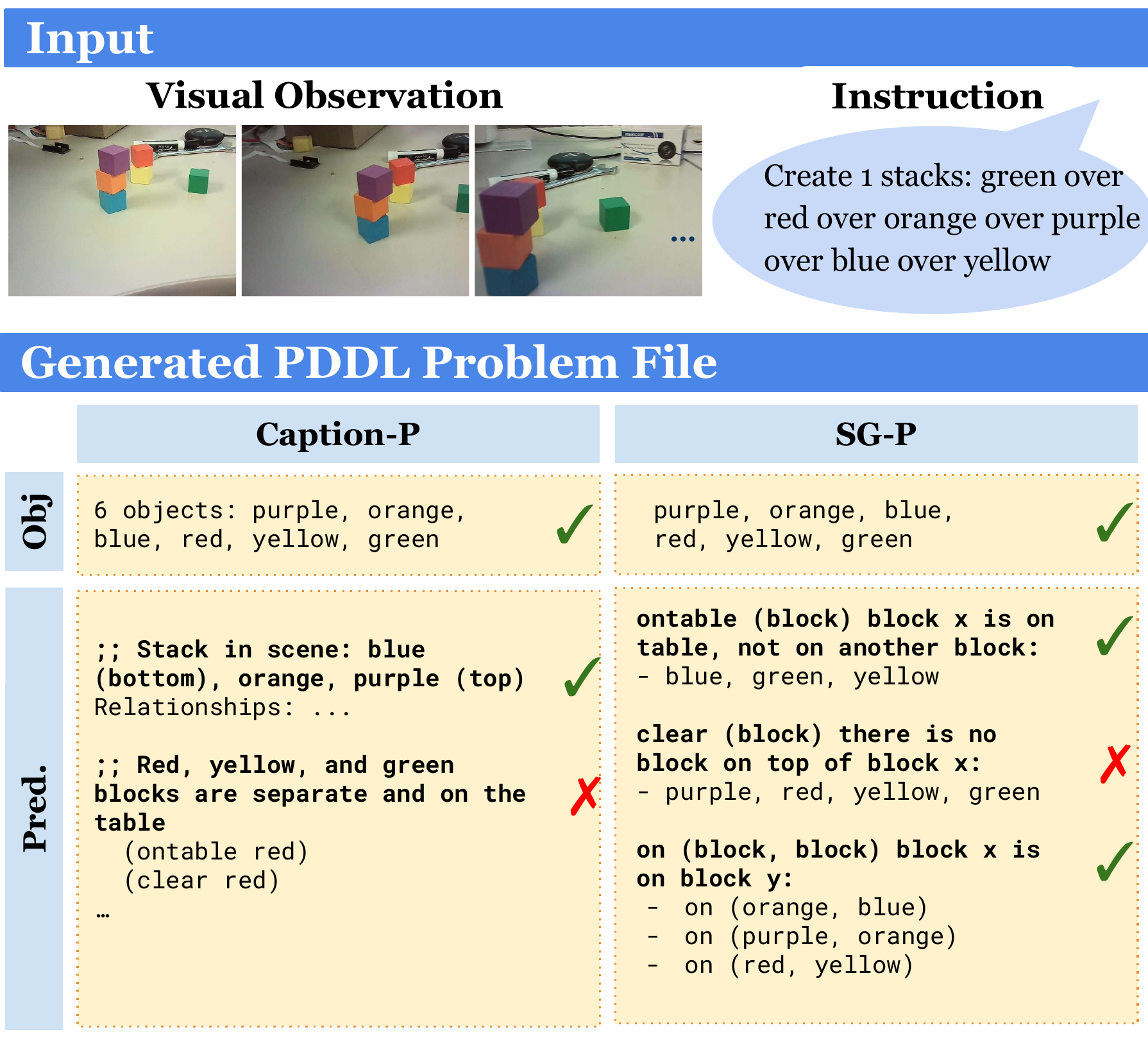}
        \caption{Examples of generated captions and scene graphs on \textsc{Blocksworld-Real}.}
        \label{fig:qualitative}
    \end{minipage}
\end{figure}

%% file: sections/related_work.tex
\section{Related Work}

\paragraph{LLM as Planners and Formalizers}
In purely textual planning, LLMs have been extensively studied both as planners \citep{wei-etal-2025-plangenllms} and as formalizers \citep{tantakoun-etal-2025-llms}. As planners, LLMs have been reported to perform strongly on short-horizon planning \citep{huang2022language, hu2023look, ahn2022can} but their performance degrades on complex, long-horizon planning tasks \citep{10.5555/3666122.3669442,valmeekam2025a}. As formalizers, LLMs have been posed to deliver increased robustness and interpretability \citep{lyu-etal-2023-faithful,zhao2023large,guan2023leveraging} though such benefit has been partially verified and partially questioned on planning tasks with increased complexity \citep{zuo-etal-2025-planetarium,huang-zhang-2025-limit,kagitha2025addressingchallengesplanninglanguage}. Regardless, studies of either method on multimodal data has been limited.

\paragraph{VLM as Planners and Formalizers}
Some preliminary steps have been taken on formalizer approaches in the vision domain~\citep{radford2021learning, huang2023voxposercomposable3dvalue, ahn2022can}, yet current success is often shown on multimodal tasks with the aid of non-vision inputs~\citep{kwon2025fastaccuratetaskplanning, li2025bilevellearningbilevelplanning, liang2025visualpredicatorlearningabstractworld}.
For tasks that focus only on vision, they rely mainly on close-vocabulary~\citep{shirai2024visionlanguageinterpreterrobottask, siburian2025groundedvisionlanguageinterpreterintegrated}, few-shot techniques~\citep{herzog2025domainconditionedscenegraphsstategrounded} in overly simplified scenarios \citep{dang2025planningvisionlanguagemodelsuse}. 
More broadly, techniques that reason over text-and-image inputs attempt to grounded object information by identifying visible objects in the current scenario~\citep{song2023llm, huang2023grounded}, and employ caption-based approaches~\citep{yang2025embodiedbench}. More advanced methods utilize structured symbolic representations, such as scene graphs in both 2D and 3D scenarios, to improve reasoning capabilities~\citep{herzog2025domain, mitra2024compositional, wang2024predicate, jiaosequential, zhu2021hierarchical}, and develop open-domain scene graph grounding models~\citep{zhang2025open, gu2024conceptgraphs}.

\paragraph{Multimodal Planning Benchmarks}
A series of works has been established for evaluating embodied agents' performance, spanning tasks from high-level representation understanding~\citep{shridhar2020alfred, li2023behavior, shirai2024vision},  to fine-grained, low-level continuous control~\citep{zheng2022vlmbench, khanna2024goat, zhang2024vlabench}. These evaluations cover diverse application domains such as object manipulation~\citep{zheng2022vlmbench, ahn2022can, valmeekam2023planbench}, household tasks~\citep{shridhar2020alfworld, liu2024visualagentbench, merler2025viplan}, and navigation scenarios~\citep{ gadre2023cows, jain2024streetnav}, progressively evolving from artificial, structured tasks toward more realistic challenges relevant to human activities. Moreover, there is an increasing shift from purely simulated environments~\citep{kolve2017ai2, szot2021habitat} toward real-world robotic deployments~\citep{ zitkovich2023rt, driess2023palm, khazatsky2024droid}. Evaluation settings have shifted also from primarily single-view, fully observable symbolic scenarios~\citep{shridhar2020alfred, valmeekam2023planbench, li2023behavior} to more complex, multi-view setups with partial or complete observability~\citep{yan2018chalet, khazatsky2024droid}, reflecting greater realism in embodied agent interactions.

\paragraph{Open-Vocabulary Detection and Scene Graph Generation} 
The ability to detect and localize objects described in natural language has become increasingly important for vision-language tasks. With an evolving level of sophistication over earlier works~\citep{kazemzadeh2014referitgame, yu2018mattnet, wu2022language}, the emergence of large-scale vision-language pretraining, such as CLIP~\citep{radford2021learning}, GLIP~\citep{li2022grounded}, and Grounding DINO~\citep{liu2024grounding}, have achieved remarkable performance.
In addition to object detection, scene graphs provide structured representations of visual scenes. Pioneering work by Lu \emph{et al.}~\citep{lu2016visual} and Xu \emph{et al.}~\citep{xu2017scene} introduced visual relationship detection through language priors, whose key challenges have been subsequently addressed ~\citep{zellers2018neural, tang2020unbiased, knyazev2020graph}. Recent advances have explored knowledge-embedded approaches~\citep{chen2019knowledge}, natural language supervision~\citep{zhong2021learning}, and temporal dynamics in video scenes~\citep{cong2021spatial}. While traditional scene graph generation focuses on single-image parsing, our work requires VLLMs to construct scene graphs from multiple partial views and translate them into formal logical representations for planning. This poses unique challenges in cross-view consistency and the integration of spatial reasoning with symbolic planning languages.

%% file: sections/conclusion.tex
\section{Conclusion}
\label{sec:conclusion}

In this work, we evaluated the effectiveness of \vaf in solving long-horizon visual-language planning tasks.
We proposed two novel benchmarks, \pddl{Blocksworld-Real} and \pddl{Alfred-Multi}, that for the first time present partially observable, multi-view environments in the Vision-PDDL-Planning domain.
Our evaluation of five \vaf pipelines demonstrates their effectiveness over the end-to-end baseline and establishes VLMs as blind thinkers whose success depends on a crucial interplay of vision and language.
Limitations in current approaches, such as token inefficiency and insufficient recall, remain to be addressed by upcoming research.